\documentclass{aiccc2026}

\usepackage{multirow}
\usepackage[normalem]{ulem}
\useunder{\uline}{\ul}{}

\begin{document}

\papertitle{RECAP: Local Hebbian Prototype Learning as a Self-Organizing Readout for Reservoir Dynamics}
\papersubtitle{}

\paperauthor{Heng Zhang*}
  {International Research Center for Neurointelligence (WPI-IRCN), UTIAS, University of Tokyo, Japan}
  {rogerzhangheng@gmail.com}

\begin{aicccabstract}
Robust perception in brains is often attributed to high-dimensional population activity together with local plasticity mechanisms that reinforce recurring structure. In contrast, most modern image recognition systems are trained by error backpropagation and end-to-end gradient optimization, which are not naturally aligned with local computation and local plasticity. We introduce \textbf{RECAP} (\textbf{R}eservoir Computing with H\textbf{E}bbian \textbf{C}o-\textbf{A}ctivation \textbf{P}rototypes), a bio-inspired learning strategy for \textit{robust image classification} that couples untrained reservoir dynamics with a self-organizing Hebbian prototype readout. RECAP discretizes time-averaged reservoir responses into activation levels, constructs a co-activation mask over reservoir unit pairs, and incrementally updates class-wise prototype matrices via a Hebbian-like potentiation--decay rule. Inference is performed by overlap-based prototype matching. The method avoids error backpropagation and is naturally compatible with online prototype updates. We illustrate the resulting robustness behavior on MNIST-C, where RECAP remains resilient under diverse corruptions without exposure to corrupted training samples.
\end{aicccabstract}

\ccsconcepts{%
\textbullet\ Computing methodologies $\rightarrow$ Neural networks;
\textbullet\ Computing methodologies $\rightarrow$ Bio-inspired approaches;
\textbullet\ Computing methodologies $\rightarrow$ Object recognition
}

\additionalkw{Reservoir computing, Hebbian learning, self-organization, prototype learning, robustness, common corruptions, MNIST-C, bio-inspired vision}

\par\vspace{7pt}%
{\fontsize{8}{12}\selectfont\noindent Accepted at the 9th Artificial Intelligence and Cloud Computing Conference (AICCC 2026), Tokyo, Japan, December 18--21, 2026.\par}

\authornote{* Corresponding author.}

\section{Introduction}
\label{sec:introduction}

Biological perception is remarkably robust: humans recognize objects under diverse degradations without explicit training on every possible distortion.
This reliability is often attributed to population-level neural representations and local plasticity mechanisms that consolidate stable structure from repeated experience \cite{hebb2005organization,caporale2008spike}.
These observations motivate a long-standing goal in neuro-inspired computation: build learning systems whose robustness emerges from \emph{dynamics and self-organization}, rather than from exhaustive optimization or exposure to every possible perturbation.

Modern machine vision frequently exhibits the opposite behavior.
Deep networks excelling on clean benchmarks can be fragile under realistic distribution shifts, including non-adversarial corruptions produced by sensors or environments \cite{hendrycks2019robustness}.
While adversarial perturbations have been studied extensively \cite{goodfellow2014explaining}, a practically important failure mode is degradation under \emph{common corruptions} such as noise, blur, weather artifacts, and digital distortions.
Moreover, standard remedies typically involve augmentation or specialized training objectives, increasing computational cost and tying robustness to assumed perturbation models. In addition, the dominant training mechanism in modern deep learning, error backpropagation, relies on non-local credit assignment and assumptions (e.g., precise feedback/weight transport) that are difficult to reconcile with known biological circuitry. This has motivated learning frameworks that emphasize local computation and local plasticity as minimal plausibility criteria \cite{lillicrap2020backpropagation,bengio2015towards}.

Reservoir computing (RC) offers a neuro-inspired alternative \cite{jaeger2001echo,maass2002real,zhang2023survey}.
In RC, an untrained recurrent network generates rich high-dimensional dynamics, confining learning to the readout layer.
This framework has shown promise in settings where robustness to noise is desirable \cite{jalalvand2018application,laje2013robust}.
However, standard linear readouts (e.g., ridge regression) can still be brittle when the representation is not linearly separable or when corruptions distort the feature geometry.

\begin{figure}[htbp]
\centering
\includegraphics[width=0.65\linewidth]{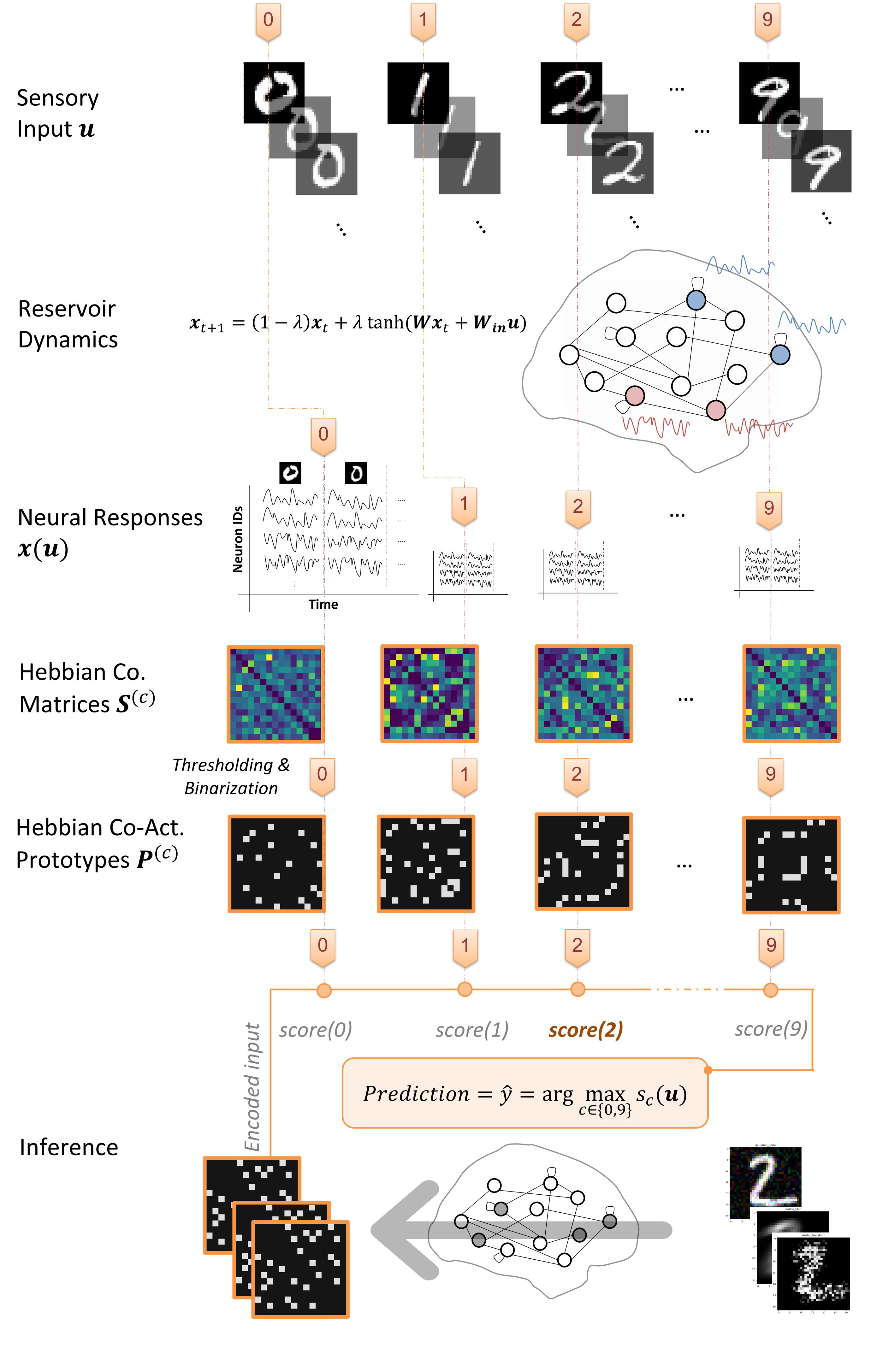}
\caption{RECAP for robust image classification. An input image drives an untrained echo-state reservoir, producing temporal population activity that is time-averaged and discretized into activation levels. The discretized population code is converted into a binary co-activation mask, which updates a class-wise continuous prototype state via a Hebbian-like potentiation--decay rule. After thresholding, inference is performed by matching the test mask against all class prototypes and selecting the highest-overlap class.}
\label{fig:overview}
\end{figure}

We propose \textbf{RECAP} (\textbf{R}eservoir Computing with H\textbf{E}bbian \textbf{C}o-\textbf{A}ctivation \textbf{P}rototypes)\footnote{Pronounced ``re-cap.''}, a self-organizing readout designed for robustness to common corruptions without training on corrupted samples.
Given an input image, the reservoir produces a temporal response that we time-average into a stable activity vector.
This vector is discretized into a small number of activation levels, yielding a \emph{co-activation mask} encoding which reservoir units share the same discrete state.
For each class, RECAP learns a sparse binary prototype updated by a simple potentiation-and-decay rule inspired by Hebbian plasticity \cite{hebb2005organization}, where backpropagation and end-to-end gradient-based optimization are not required.
At inference, classification reduces to template matching: the predicted label corresponds to the prototype with highest overlap.
Because the prototype updates are incremental, the readout can in principle be updated online as new data arrive, suggesting a potential bridge to adaptive and continual learning settings.
Figure~\ref{fig:overview} provides a graphical overview of RECAP, showing how untrained reservoir dynamics are converted into co-activation masks and then matched against class-wise prototypes for inference.

We evaluate RECAP on \textbf{MNIST-C}, adapting ImageNet-C corruption types to MNIST \cite{hendrycks2019robustness}.
Across 15 corruption types at five severity levels, RECAP achieves relative mCE of \textbf{34.1\%}, compared with \textbf{52.1\%} for MLP, \textbf{55.0\%} for ESN-Ridge (ablation), and near-baseline performance for ResNet-18 and AlexNet---despite never observing corrupted images during training.
This gain involves a tradeoff in clean accuracy, which we discuss explicitly: our goal is not to outperform optimized deep architectures on clean data, but to demonstrate that a biologically inspired, self-organizing readout can provide strong corruption robustness under constrained training.

\subsection{Contributions}
\begin{itemize}
    \item We introduce \textbf{RECAP}, a bio-inspired learning strategy that combines untrained reservoir dynamics with a Hebbian co-activation prototype readout trained without error backpropagation or gradient descent.
    \item We show that RECAP yields strong robustness to common corruptions on MNIST-C \emph{without training on corrupted data}: it achieves the lowest Relative mCE of all compared models (34.1\% vs.\ 52.1\% for MLP and 99.9\% for ResNet-18), at the cost of a higher clean error (11.7\%).
    \item An ablation on the same fixed reservoir shows that the readout learning rule drives this robustness: replacing the ridge readout (ESN-Ridge) with RECAP reduces Relative mCE from 55.0\% to 34.1\%.
    \item We discuss how the local, incremental prototype updates make RECAP naturally compatible with online adaptation.
\end{itemize}

\section{Related Work}
\label{sec:related}

\subsection{Robustness Under Common Corruptions}
\begin{figure}[ht]
    \centering
    \includegraphics[width=0.95\linewidth]{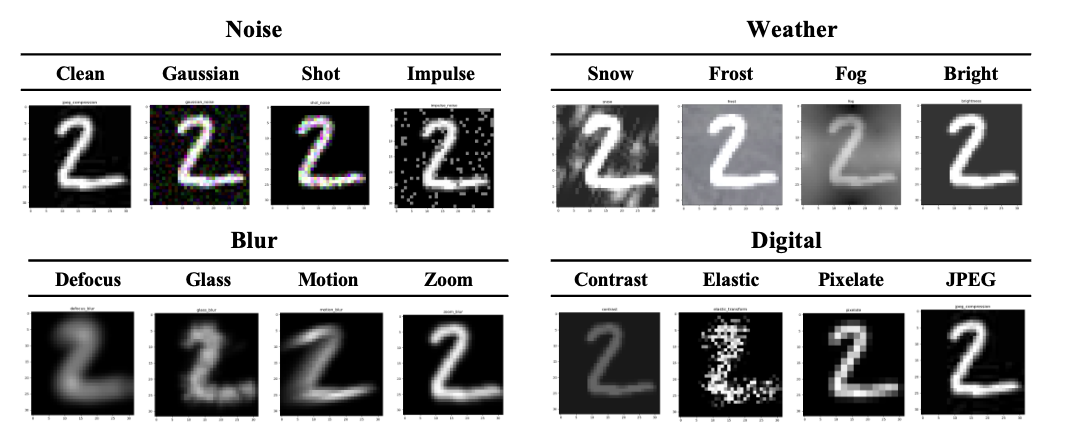}
    \caption{Corrupted examples of the MNIST-C dataset used in the experiments. The method consists of 15 types of algorithmically generated corruptions from noise (\textbf{top-left}), blur (\textbf{bottom-left}), weather (\textbf{top-right}), and digital categories (\textbf{bottom-right}). Each type has 5 levels of severity (i.e., 75 distinct corruptions).}
    \label{fig:corrupted_imagenet_mnist}
\end{figure}

Strong clean-data performance does not guarantee reliable behavior under realistic perturbations. Beyond adversarial examples crafted to induce errors \cite{goodfellow2014explaining,madry2017towards}, vision systems must handle \emph{natural corruptions}. Hendrycks and Dietterich introduced standardized benchmarks (e.g., ImageNet-C) with metrics including mean Corruption Error (mCE) to quantify how performance degrades with corruption severity \cite{hendrycks2019robustness}. In this work, we adapt this benchmark to the MNIST test set, producing \textbf{MNIST-C} (see Figure~\ref{fig:corrupted_imagenet_mnist}).

The gap between benchmark accuracy and real-world reliability extends beyond explicit corruptions. Subtle image transformations substantially impact convolutional networks \cite{azulay2019deep}, and evaluation on newly collected test sets reveals non-trivial generalization gaps \cite{recht2018cifar}. Moreover, robustness and accuracy objectives often exhibit tension \cite{tsipras2018robustness}, suggesting that robustness may require different training signals, model biases, or representational strategies. Finally, robustness claims benefit from careful evaluation protocols, as defenses can exploit weaknesses in testing procedures \cite{athalye2018obfuscated,uesato2018adversarial}.

\subsection{Reservoir Computing}
Reservoir computing (RC) offers a neuro-inspired framework in which a fixed, randomly connected recurrent network transforms inputs into high-dimensional dynamical states, with learning restricted to a simple readout. Echo State Networks (ESNs) \cite{jaeger2001echo} and Liquid State Machines (LSMs) \cite{maass2002real} are foundational instances of this paradigm, established as practical approaches for temporal processing and nonlinear system modeling \cite{lukovsevivcius2009reservoir,lukovsevivcius2012practical}. More recently, RC has expanded to physical and neuromorphic substrates, emphasizing efficient computation through dynamical richness rather than end-to-end optimization \cite{tanaka2019recent}.

In standard RC, the readout is linear and trained via ridge regression \cite{hoerl1970ridge}, leveraging the reservoir's nonlinear expansion for separability. However, this places a strong burden on the readout assumption: if the representation is not linearly separable or test-time shifts distort feature geometry, linear readouts become brittle. This motivates exploration of alternative readout strategies, including biologically inspired adaptation rules such as intrinsic plasticity \cite{steil2007online,schrauwen2008improving}.

\begin{figure}[ht]
\centering
\includegraphics[width=0.99\linewidth]{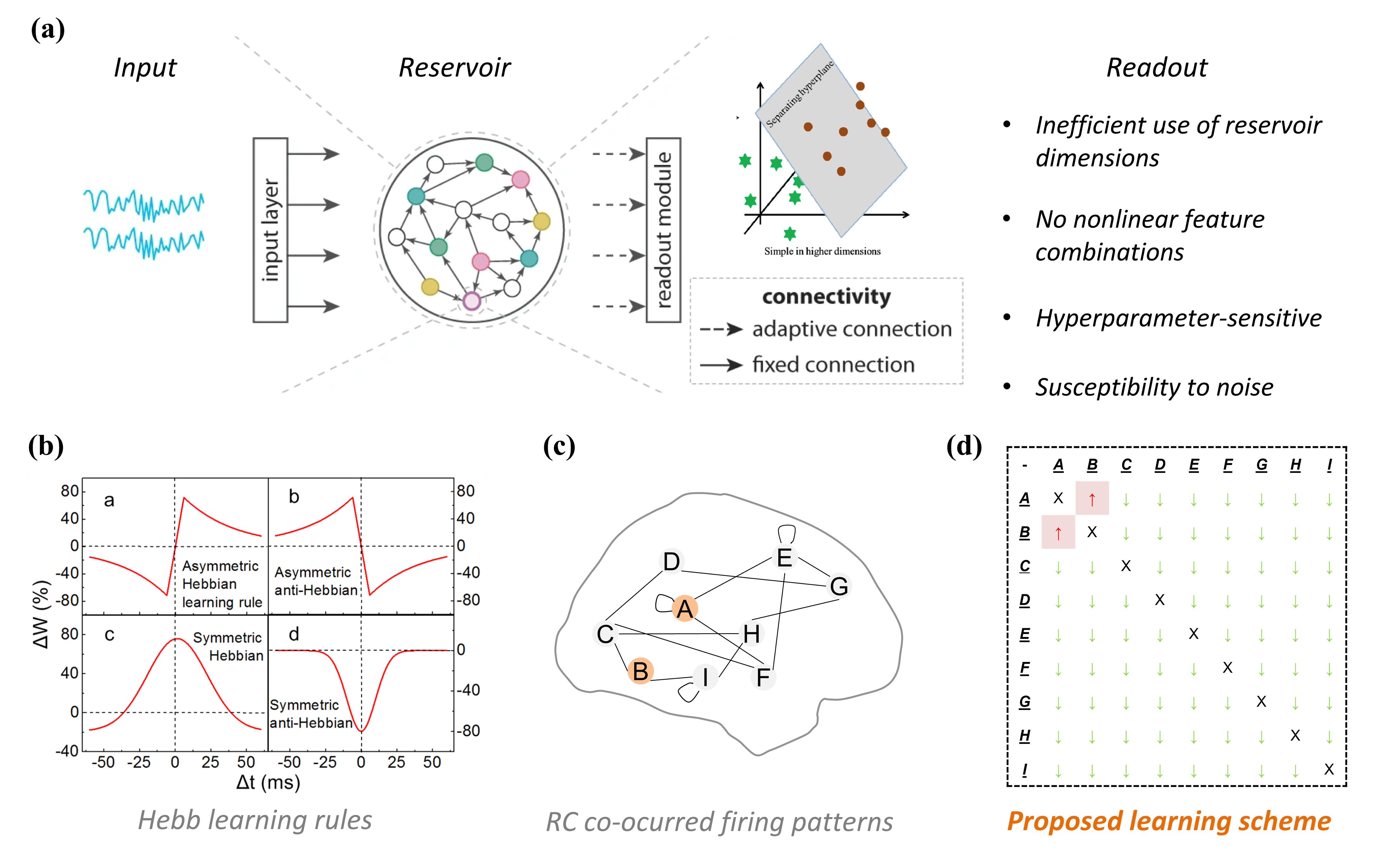}
\caption{Motivation and intuition for the RECAP readout. (a) Standard reservoir computing relies on a typically linear readout, which can be sensitive to feature distortions under input degradations. (b) Local Hebbian-style plasticity provides an inspiration for strengthening repeatedly co-activated relations \cite{li2014activity}. (c) Repeated inputs can induce structured co-activation patterns in reservoir population responses. (d) RECAP leverages these ideas by learning class-wise relational prototypes from co-activation masks using a Hebbian-like potentiation--decay update and performing inference via prototype matching. Figures modified from \cite{li2014activity,suarez2024connectome,fusi2016neurons}.}
\label{fig:rc_hebb_related_work}
\end{figure}

Although RC is often associated with temporal tasks, it has been applied to vision, including digit recognition \cite{schaetti2016echo} and recognition under noisy conditions \cite{jalalvand2018application}. However, most vision-oriented RC work evaluates on clean test sets rather than systematic corruption robustness. Our approach directly targets robustness to common corruptions \cite{hendrycks2019robustness} while keeping recurrent dynamics untrained and concentrating novelty on a Hebbian readout strategy.
Figure~\ref{fig:rc_hebb_related_work} summarizes the motivation for moving beyond linear readouts in RC and illustrates the intuition behind our prototype-based Hebbian readout.

\subsection{Biological Plausibility and Local Learning Rules}
\label{sec:bio_plausibility_optional}
Backpropagation is the dominant training algorithm in modern deep learning, but its biological plausibility has been questioned because it relies on non-local credit assignment \cite{lillicrap2020backpropagation,bengio2015towards}.
This has motivated alternative learning frameworks that approximate or replace backprop using local computation and local plasticity, including predictive-coding formulations with local Hebbian updates \cite{tscshantz2023hybrid,oliviers2025bidirectional}.
Our work follows this direction in a limited sense: learning is performed by local potentiation--decay updates on co-activation structure, without explicit backpropagation.
Our claims of biological plausibility are therefore restricted to these principles (local computation and local plasticity), and we do not claim to be more (or less) biologically plausible than other models that satisfy the same criteria.

\subsection{Self-Organization and Dynamical Inspirations}
Self-organization in nonlinear dynamical systems offers a complementary perspective on neuro-inspired learning, emphasizing how structured representations can emerge from simple update rules and iterative dynamics \cite{strogatz2018nonlinear,ravichandran2025unsupervised}. A recently proposed model called SyncMap exemplifies this approach: a self-organizing nonlinear dynamical system that discovers temporal association structure through local dynamics and downstream clustering \cite{vargas2021syncmap,zhang2023symmetrical,mao2023magnum}.

Our work draws high-level inspiration from such dynamical approaches. While SyncMap was designed for unsupervised structure discovery in temporal sequences, we adapt the principle of self-organizing correlation learning to the supervised setting of robust image classification.

\section{Methodology}
\label{sec:method}

\begin{figure}[ht]
\centering
\includegraphics[width=0.80\linewidth]{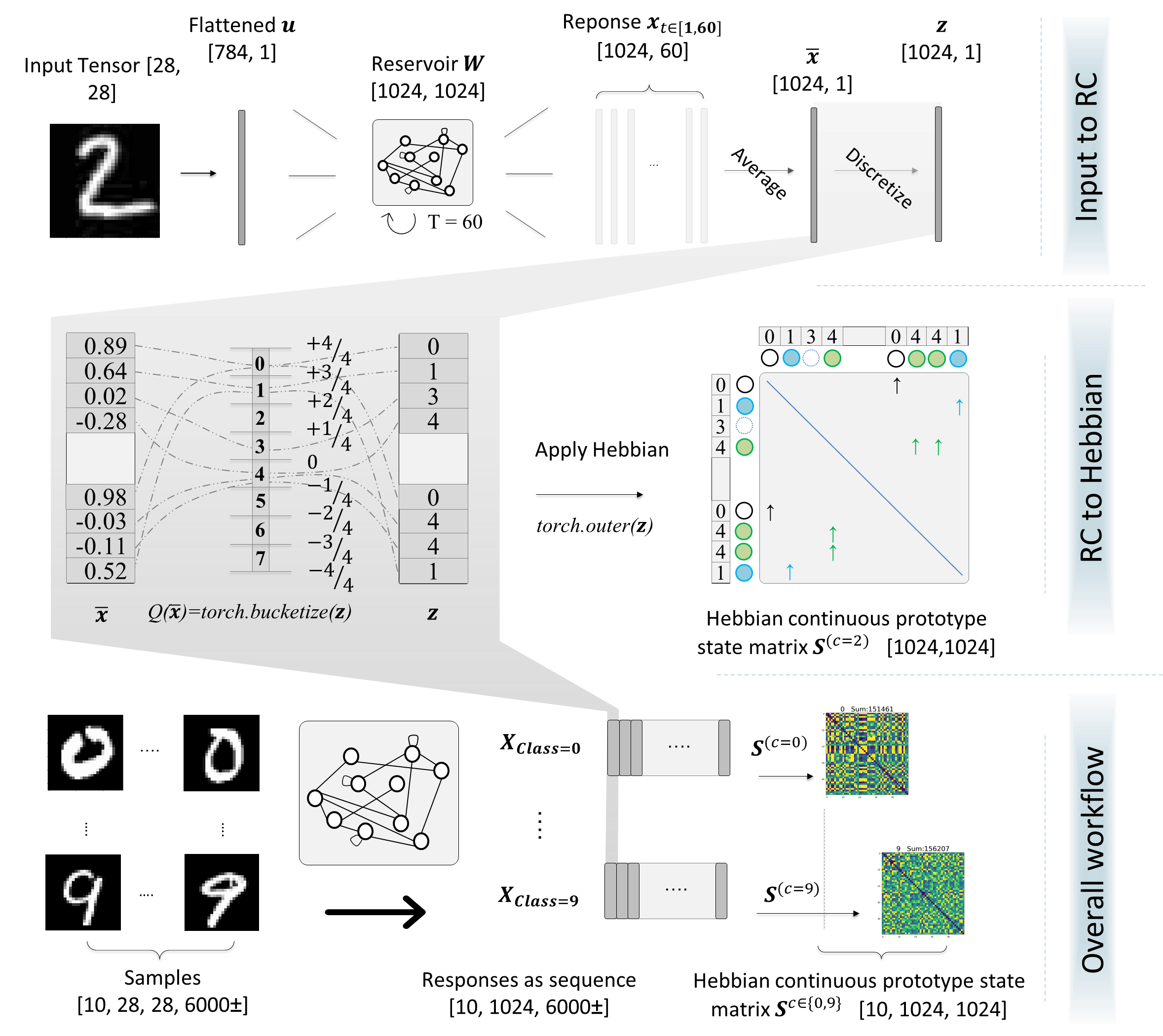}
\caption{Workflow for prototype formation. An input image is processed by the reservoir for $T$ steps, yielding a time-averaged state vector that is quantized into $K$ discrete levels. This induces a binary co-activation mask over unit pairs. During training, each sample updates its class prototype via Hebbian potentiation--decay; prototypes are then thresholded into binary templates for inference.}
\label{fig:method}
\end{figure}

RECAP combines (i) an \emph{untrained} recurrent reservoir that transforms an input image into a high-dimensional population response, with (ii) a \emph{self-organizing} readout that learns class-wise relational prototypes via a simple Hebbian-like potentiation--decay rule.
This section presents the full method with a unified mathematical formulation. Figure~\ref{fig:method} illustrates the end-to-end workflow of RECAP, including reservoir feature generation, discretization into activation levels, construction of the co-activation mask, and the Hebbian prototype update used during training.

\subsection{Notation and Overview}
Let $\mathcal{D}=\{(\mathbf{u}^{(n)},y^{(n)})\}_{n=1}^{N}$ be a labeled dataset with $C$ classes, $y^{(n)}\in\{1,\dots,C\}$.
An input image is flattened and normalized to $\mathbf{u}\in\mathbb{R}^{d}$.
The reservoir contains $N_r$ recurrent units; its state at discrete time $t$ is $\mathbf{x}_t\in\mathbb{R}^{N_r}$.
RECAP constructs, for each class $c$, a binary prototype matrix $\mathbf{P}^{(c)}\in\{0,1\}^{N_r\times N_r}$ that encodes stable co-activation relations among reservoir units for that class. Two intermediate objects are used:
\begin{itemize}
    \item A \emph{co-activation mask} $\mathbf{M}(\mathbf{u})\in\{0,1\}^{N_r\times N_r}$ derived from a discretized reservoir response to $\mathbf{u}$.
    \item A \emph{continuous} prototype state $\mathbf{S}^{(c)}\in[0,1]^{N_r\times N_r}$ updated online during training and later sparsified/binarized into $\mathbf{P}^{(c)}$.
\end{itemize}

\noindent We use $\odot$ to denote element-wise multiplication, $\mathbf{1}[\cdot]$ the indicator function, and $\mathrm{clip}_{[0,1]}(\cdot)$ the element-wise clamp into $[0,1]$.

\subsection{Untrained Reservoir Feature Generator}
\label{subsec:rc}

We adopt a standard leaky echo-state network \cite{jaeger2001echo,lukovsevivcius2012practical}.
Given a constant input $\mathbf{u}$, the reservoir evolves for $T$ cycles. $T$ is set long enough to wash out the initial transient states:
\begin{equation}
\mathbf{x}_{t+1}
=
(1-\lambda)\mathbf{x}_t
+
\lambda\,\tanh\!\left(
\mathbf{W}\mathbf{x}_t + \mathbf{W}_{\mathrm{in}}\mathbf{u}
\right),
\qquad t=0,\dots,T-1,
\label{eq:rc_update}
\end{equation}
where $\mathbf{W}\in\mathbb{R}^{N_r\times N_r}$ is a sparse recurrent weight matrix, $\mathbf{W}_{\mathrm{in}}\in\mathbb{R}^{N_r\times d}$ is an input weight matrix, and $\lambda\in(0,1]$ is a leaking rate.
In RECAP the reservoir is \emph{not} trained and is randomly initialized and kept fixed.
To obtain a stable representation of a static image, we repeatedly inject the same $\mathbf{u}$ for $T$ cycles and compute a time-averaged state:
\begin{equation}
\bar{\mathbf{x}}(\mathbf{u})
=
\frac{1}{T}\sum_{t=1}^{T}\mathbf{x}_t
\in\mathbb{R}^{N_r}.
\label{eq:rc_avg}
\end{equation}
This average reduces sensitivity to transient dynamics and provides a single vector representation per image.

\subsection{Discretization and Co-Activation Mask Construction}
\label{subsec:mask}

RECAP converts $\bar{\mathbf{x}}(\mathbf{u})$ into a discrete code that is less sensitive to small amplitude perturbations.
Let $K$ be the number of discretization levels (we use a small fixed $K$, e.g., $K=8$ in our experiments).
Define a scalar quantizer $Q:\mathbb{R}\rightarrow\{0,\dots,K-1\}$ using fixed bin edges $\{b_k\}_{k=0}^{K}$ with $b_0=-1$ and $b_K=1$.
With uniform bins, one convenient choice is
\begin{equation}
b_k = -1 + \frac{2k}{K},\qquad k=0,\dots,K.
\label{eq:bin_edges}
\end{equation}
Each reservoir unit is assigned a discrete level
\begin{equation}
z_i(\mathbf{u}) = Q(\bar{x}_i(\mathbf{u})),\qquad i=1,\dots,N_r,
\label{eq:quantize}
\end{equation}
yielding a code vector $\mathbf{z}(\mathbf{u})\in\{0,\dots,K-1\}^{N_r}$.

We then define a binary \emph{co-activation mask} $\mathbf{M}(\mathbf{u})$ over unit pairs:
\begin{equation}
M_{ij}(\mathbf{u})
=
\mathbf{1}\!\left[z_i(\mathbf{u})=z_j(\mathbf{u})\right]\cdot \mathbf{1}[i\neq j],
\qquad 1\le i,j\le N_r.
\label{eq:mask_def}
\end{equation}
Intuitively, $M_{ij}=1$ indicates that units $i$ and $j$ occupy the same discrete activation level under the current input, forming a relational, population-level signature of the reservoir response.
The diagonal is always excluded ($M_{ii}=0$), preventing self-relations from dominating downstream matching.

\subsection{Hebbian Co-Activation Prototype Dynamics}
\label{subsec:hecap_dynamics}

For each class $c\in\{1,\dots,C\}$, RECAP maintains a continuous prototype state
$\mathbf{S}^{(c)}\in[0,1]^{N_r\times N_r}$ initialized at zero:
\begin{equation}
\mathbf{S}^{(c)} \leftarrow \mathbf{0}.
\label{eq:init_S}
\end{equation}
Training is \emph{class-conditional}: $\mathbf{S}^{(c)}$ is updated only using samples with label $c$ (i.e., ``positive-only'' updates per class).
This yields $C$ independently learned prototype states.

Let $\mathbf{M}^{(n)}=\mathbf{M}(\mathbf{u}^{(n)})$ be the mask for a training sample with label $y^{(n)}=c$.
We update $\mathbf{S}^{(c)}$ using a simple potentiation--decay rule inspired by local plasticity \cite{hebb2005organization,caporale2008spike} and self-organizing attraction/repulsion dynamics (in spirit) \cite{vargas2021syncmap}.
For each pair $(i,j)$:
\begin{equation}
S^{(c)}_{ij}
\leftarrow
\begin{cases}
\mathrm{clip}_{[0,1]}\!\left(S^{(c)}_{ij} + \eta_{+}\right), & \text{if } M^{(n)}_{ij}=1,\\[4pt]
\eta_{-}\, S^{(c)}_{ij}, & \text{if } M^{(n)}_{ij}=0,
\end{cases}
\label{eq:update_piecewise}
\end{equation}
where $\eta_{+}>0$ is the potentiation rate and $\eta_{-}\in(0,1)$ is the decay rate.

Equation~\eqref{eq:update_piecewise} is a direct and unambiguous combination of the two intended operations:
(1) \emph{positive update} (addition) on co-activated pairs, and
(2) \emph{negative update} (multiplicative decay) on non-co-activated pairs.
In matrix form, the same update can be written compactly as
\begin{equation}
\mathbf{S}^{(c)}
\leftarrow
\mathrm{clip}_{[0,1]}\!\Big(
\mathbf{M}^{(n)} \odot (\mathbf{S}^{(c)} + \eta_{+}\mathbf{J})
\;+\;
(\mathbf{J}-\mathbf{M}^{(n)}) \odot (\eta_{-}\mathbf{S}^{(c)})
\Big),
\label{eq:update_matrix}
\end{equation}
where $\mathbf{J}$ denotes the all-ones matrix.
After each update we enforce the zero diagonal:
\begin{equation}
S^{(c)}_{ii} \leftarrow 0,\qquad i=1,\dots,N_r.
\label{eq:zero_diag}
\end{equation}

\paragraph{Interpretation.}
As in Hebbian learning, pairs that repeatedly fall into the same discrete activation level across many class-$c$ examples are incrementally strengthened (potentiation) and saturate due to clamping.
Pairs that are not consistently co-activated gradually fade (decay).
This simple dynamic promotes stable, class-specific relational templates without gradient-based optimization.

\subsection{Prototype Binarization}
\label{subsec:normalize}

Because different classes can induce different overall scales in $\mathbf{S}^{(c)}$, RECAP normalizes prototypes by enforcing equal density across classes before comparison.
Given a target sparsity fraction $p\in(0,1)$ (e.g., $p=0.30$ in our experiments), we compute a class-specific threshold $\theta_c$ as the $(1-p)$ quantile of off-diagonal entries:
\begin{equation}
\theta_c
=
\mathrm{Quantile}_{1-p}\left(\left\{S^{(c)}_{ij}\,:\, i\neq j\right\}\right).
\label{eq:theta}
\end{equation}
The final binary co-activation prototype is
\begin{equation}
P^{(c)}_{ij}
=
\mathbf{1}\!\left[S^{(c)}_{ij}\ge \theta_c\right]\cdot \mathbf{1}[i\neq j],
\qquad 1\le i,j\le N_r.
\label{eq:binarize}
\end{equation}
This step ensures that each $\mathbf{P}^{(c)}$ contains approximately the same fraction of ones, making overlap scores comparable across classes.

\subsection{Inference by Prototype Matching}
\label{subsec:inference}

For a test image $\mathbf{u}$, we compute its mask $\mathbf{M}(\mathbf{u})$ via
\eqref{eq:rc_update}--\eqref{eq:mask_def}.
We then compute a class score using overlap (logical-AND followed by summation):
\begin{equation}
s_c(\mathbf{u})
=
\sum_{i\neq j} P^{(c)}_{ij}\, M_{ij}(\mathbf{u})
=
\left\langle \mathbf{P}^{(c)},\mathbf{M}(\mathbf{u}) \right\rangle_F,
\label{eq:score}
\end{equation}
where $\langle\cdot,\cdot\rangle_F$ denotes the Frobenius inner product.
Finally, prediction is obtained by
\begin{equation}
\hat{y}(\mathbf{u}) = \arg\max_{c\in\{1,\dots,C\}} s_c(\mathbf{u}).
\label{eq:predict}
\end{equation}

\subsection{Role of the Readout Representation}
\label{subsec:readout_role}
The reservoir in RECAP is random, fixed, and shared with the ESN-Ridge baseline; RECAP changes only the readout representation, i.e., how the reservoir response is encoded, learned, and compared. Each stage is designed to tolerate input corruptions:
(i) \emph{discretization} (Eq.~\eqref{eq:quantize}) leaves a unit's code unchanged unless a perturbation moves it across a bin edge;
(ii) the \emph{co-activation mask} (Eq.~\eqref{eq:mask_def}) records whether two units share a level, not their exact values, capturing relational structure rather than magnitude;
(iii) the \emph{potentiation--decay rule} (Eq.~\eqref{eq:update_matrix}) keeps relations that recur across a class and lets sample-specific ones fade; and
(iv) \emph{overlap matching} (Eq.~\eqref{eq:score}) pools evidence over many unit pairs, so flipping a fraction of mask entries only partially lowers the score of the correct class.
A linear readout instead weights continuous activations directly, so amplitude distortions propagate to its decision.
Section~\ref{sec:results} tests this by comparing RECAP with ESN-Ridge on the same reservoir.

\section{Experiments}
\label{sec:experiments}

This section summarizes the evaluation protocol used to assess robustness under \emph{common corruptions} (see Table~\ref{tab:exp_setup}). All models are trained \emph{only} on clean MNIST training data; corrupted data are used exclusively for testing.

\begin{table}[t]
\centering
\caption{Summary of the experimental setup.}
\label{tab:exp_setup}
\resizebox{0.85\linewidth}{!}{%
\begin{tabular}{p{0.26\linewidth} p{0.68\linewidth}}
\hline
Training data & Clean MNIST training set (no corruptions).\\
Test data & Clean MNIST test set and MNIST-C: 15 corruption types $\times$ 5 severity levels.\\
Corruption families & Noise, Blur, Weather, Digital (as in ImageNet-C) \cite{hendrycks2019robustness}.\\
Baselines & MLP, ResNet-18 \cite{he2016deep}, AlexNet \cite{krizhevsky2012imagenet}, ESN-Ridge \cite{jaeger2001echo,hoerl1970ridge}.\\
Metrics & Clean error; per-corruption Relative CE; Relative mCE (mean over corruptions) \cite{hendrycks2019robustness}.\\
\hline
\end{tabular}
}
\end{table}

\subsection{MNIST-C: Common Corruptions on MNIST}
\label{subsec:mnistc}

To evaluate robustness beyond clean accuracy, we adapt the corruption methodology from ImageNet-C \cite{hendrycks2019robustness} to the MNIST test set, producing \textbf{MNIST-C}.
MNIST-C contains 15 corruption types organized into four families, each instantiated at five severity levels ($s\in\{1,\dots,5\}$), yielding 75 corrupted test sets in addition to the clean test set:
\textbf{Noise:} Gaussian noise, shot noise, impulse noise.
\textbf{Blur:} defocus blur, glass blur, motion blur, zoom blur.
\textbf{Weather:} snow, frost, fog, brightness.
\textbf{Digital:} contrast, elastic transform, pixelate, JPEG compression.
The corruption operators and severity scaling follow the standard definitions in the common-corruptions benchmark \cite{hendrycks2019robustness}.

\subsection{Training Protocol and Models}
\label{subsec:protocol_models}

\paragraph{Clean-only training.}
All models (including \textbf{RECAP}) are trained using \emph{only} the clean MNIST training set.
No model is trained or fine-tuned on corrupted images.
Robustness is evaluated by testing the trained models on the clean MNIST test set and on MNIST-C.

\paragraph{Baselines.}
We compare against four baselines shown in Table~\ref{tab:all_params}: \textbf{MLP:} a standard multilayer perceptron classifier trained with cross-entropy on clean MNIST. \textbf{ResNet-18:} a convolutional residual network adapted for MNIST classification \cite{he2016deep}.
\textbf{AlexNet:} a convolutional network used as the normalization reference for corruption-error metrics in common-corruptions evaluation \cite{krizhevsky2012imagenet,hendrycks2019robustness}.
\textbf{ESN-Ridge:} an echo state network with a ridge-regression readout, representing a conventional linear RC baseline \cite{jaeger2001echo,hoerl1970ridge}.

\begin{table}[ht]
\centering
\caption{Hyperparameters for RECAP (proposed) and baseline models.}
\label{tab:all_params}
\resizebox{0.6\linewidth}{!}{%
\begin{tabular}{lcrc}
\toprule
\textbf{Model} & \textbf{Component} & \textbf{Parameter} & \textbf{Value} \\
\midrule
\multirow{9}{*}{RECAP}
    & \multirow{3}{*}{Reservoir}
        & Number of neurons ($N$) & 1024 \\
    &   & Spectral radius ($\rho$) & 1.0 \\
    &   & Leaking rate ($\lambda$) & 0.5 \\
    &   & Sparsity & 0.9 \\
    \cmidrule{2-4}
    & Dynamics & Time steps ($T$) & 60 \\
    \cmidrule{2-4}
    & \multirow{3}{*}{Hebbian Learning}
        & Potentiation rate ($\eta_{+}$) & 0.6 \\
    &   & Decay rate ($\eta_{-}$) & 0.96 \\
    &   & Discretization levels ($K$) & 8 \\
    \cmidrule{2-4}
    & Prototype & Sparsity fraction ($p$) & 0.3 \\
\midrule
\multirow{2}{*}{ESN-Ridge}
    & Reservoir & Same as RECAP  & - \\
    \cmidrule{2-4}
    & Readout & Regularization ($\beta$) & $10^{-5}$ \\
\midrule
\multirow{4}{*}{MLP}
    & \multirow{2}{*}{Architecture}
        & Hidden neurons & 500 \\
    &   & Activation & ReLU \\
    \cmidrule{2-4}
    & \multirow{2}{*}{Training}
        & Epochs / Batch size & 20 / 100 \\
    &   & Learning rate & 0.001 \\
\midrule
\multirow{2}{*}{ResNet18}
    & \multirow{2}{*}{Architecture}
        & Upsampling filters & 64 (7$\times$7, stride 2) \\
    &   & Dropout & 0.2 \\
\midrule
\multirow{3}{*}{AlexNet}
    & \multirow{2}{*}{Adaptation}
        & Upsample layers & 2 transposed conv \\
    &   & Frozen layers & All conv layers \\
    \cmidrule{2-4}
    & Pretrained & PyTorch's model hub & ImageNet (v0.10.0) \\
\bottomrule
\end{tabular}
}
\end{table}

\subsection{Metrics}
\label{subsec:metrics}
We report clean test error and corruption robustness using Relative mCE and per-corruption Relative CE, normalized by AlexNet as in the common-corruptions protocol \cite{hendrycks2019robustness}.
Let $E^{\mathrm{Net}}_{c,s}$ be the error of model \emph{Net} on corruption type $c$ at severity $s\in\{1,\dots,5\}$ and $E^{\mathrm{Net}}_{\mathrm{clean}}$ the error on the clean test set.
The (normalized) Corruption Error and Relative Corruption Error are
\begin{equation}
\mathrm{CE}^{\mathrm{Net}}_{c}
=
\frac{\sum_{s=1}^{5} E^{\mathrm{Net}}_{c,s}}{\sum_{s=1}^{5} E^{\mathrm{AlexNet}}_{c,s}},
\qquad
\mathrm{RelCE}^{\mathrm{Net}}_{c}
=
\frac{\sum_{s=1}^{5}(E^{\mathrm{Net}}_{c,s}-E^{\mathrm{Net}}_{\mathrm{clean}})}{\sum_{s=1}^{5}(E^{\mathrm{AlexNet}}_{c,s}-E^{\mathrm{AlexNet}}_{\mathrm{clean}})}.
\label{eq:relce}
\end{equation}
Relative mCE is the mean of $\mathrm{RelCE}^{\mathrm{Net}}_{c}$ over the 15 corruption types.

\section{Results}
\label{sec:results}

Tables~\ref{tab:mnist_rmce_1} and \ref{tab:mnist_rmce_2} summarize the main results.
Although deep CNN baselines (e.g., ResNet-18) achieve strong clean accuracy, their Relative mCE remains close to the AlexNet-normalized reference level under corruptions.
In contrast, \textbf{RECAP} achieves the lowest Relative mCE (\textbf{34.1\%}), indicating substantially improved robustness across corruption families, at the cost of higher clean error.
Compared with the RC baseline using a linear ridge readout (ESN-Ridge), RECAP improves Relative mCE from 55.0\% to 34.1\%, highlighting that the robustness gain primarily comes from the \emph{RECAP readout strategy} rather than from training the recurrent dynamics.

\begin{table}[ht]
\centering
\caption[Relative mCE: Part 1]{Clean error (\%) on clean MNIST and robustness scores on MNIST-C.
We report Relative mCE (\%) and per-corruption Relative CE (\%), normalized by AlexNet as in \cite{hendrycks2019robustness} (AlexNet = 100\% by construction; lower is better).
Negative Relative CE values can occur when the average corrupted error is slightly lower than the clean error (Eq.~\eqref{eq:relce}).}

\label{tab:mnist_rmce_1}
\resizebox{0.9\textwidth}{!}{%
\begin{tabular}{lrr|rrr|rrrr}
                 & \multicolumn{1}{l}{} & \multicolumn{1}{l|}{} & \multicolumn{3}{c|}{Noise} & \multicolumn{4}{c}{Blur} \\ \hline
Network &
  \begin{tabular}[c]{@{}r@{}}Clean\\ Error\end{tabular} &
  \textbf{\begin{tabular}[c]{@{}r@{}}Relative\\ mCE\end{tabular}} &
  Gauss. &
  Shot &
  Impulse &
  Defocus &
  Glass &
  Motion &
  Zoom \\ \hline
AlexNet          & 2.1                  & 100.0                 & 100     & 100     & 100    & 100  & 100  & 100  & 100 \\
MLP              & 1.9                  & 52.1                  & 41      & 10      & 29     & 56   & 57   & 72   & 20  \\
ResNet18         & 0.9                  & 99.9                  & 56      & 77      & 31     & 114  & 123  & 123  & 132 \\
ESN-Ridge        & 6.3                  & 55.0                  & 44      & 14      & 44     & 58   & 60   & 81   & 41  \\
 \textbf{RECAP} & 11.7                 & {\ul \textbf{34.1}}   & 10      & 5       & 11     & 49   & 56   & 72   & 27 
\end{tabular}%
}
\end{table}
\begin{table}[ht]
\centering
\caption[Relative mCE: Part 2]{(Part 2) Weather, and Digital columns (\%).}
\label{tab:mnist_rmce_2}
\resizebox{0.9\textwidth}{!}{%
\begin{tabular}{lr|rrrr|rrrr}
                 & \multicolumn{1}{l|}{} & \multicolumn{4}{c|}{Weather} & \multicolumn{4}{c}{Digital} \\ \hline
Network & \textbf{\begin{tabular}[c]{@{}r@{}}Relative\\ mCE\end{tabular}} & Snow & Frost & Fog & Bright & Contrast & Elastic & Pixelate & JPEG \\ \hline
AlexNet          & 100.0                 & 100   & 100   & 100   & 100  & 100   & 100   & 100  & 100  \\
MLP              & 52.1                  & 82    & 93    & 98    & 100  & 78    & 30    & 6    & 5    \\
ResNet18         & 99.9                  & 78    & 95    & 108   & 100  & 116   & 93    & 118  & 127  \\
ESN-Ridge        & 55.0                  & 83    & 92    & 93    & 99   & 77    & 38    & 7    & -13  \\
 \textbf{RECAP} & {\ul \textbf{34.1}}   & 36    & 53    & 70    & 52   & 46    & 28    & 5    & -15 
\end{tabular}%
}
\end{table}

\begin{figure}[ht]
\centering
\includegraphics[width=0.90\linewidth]{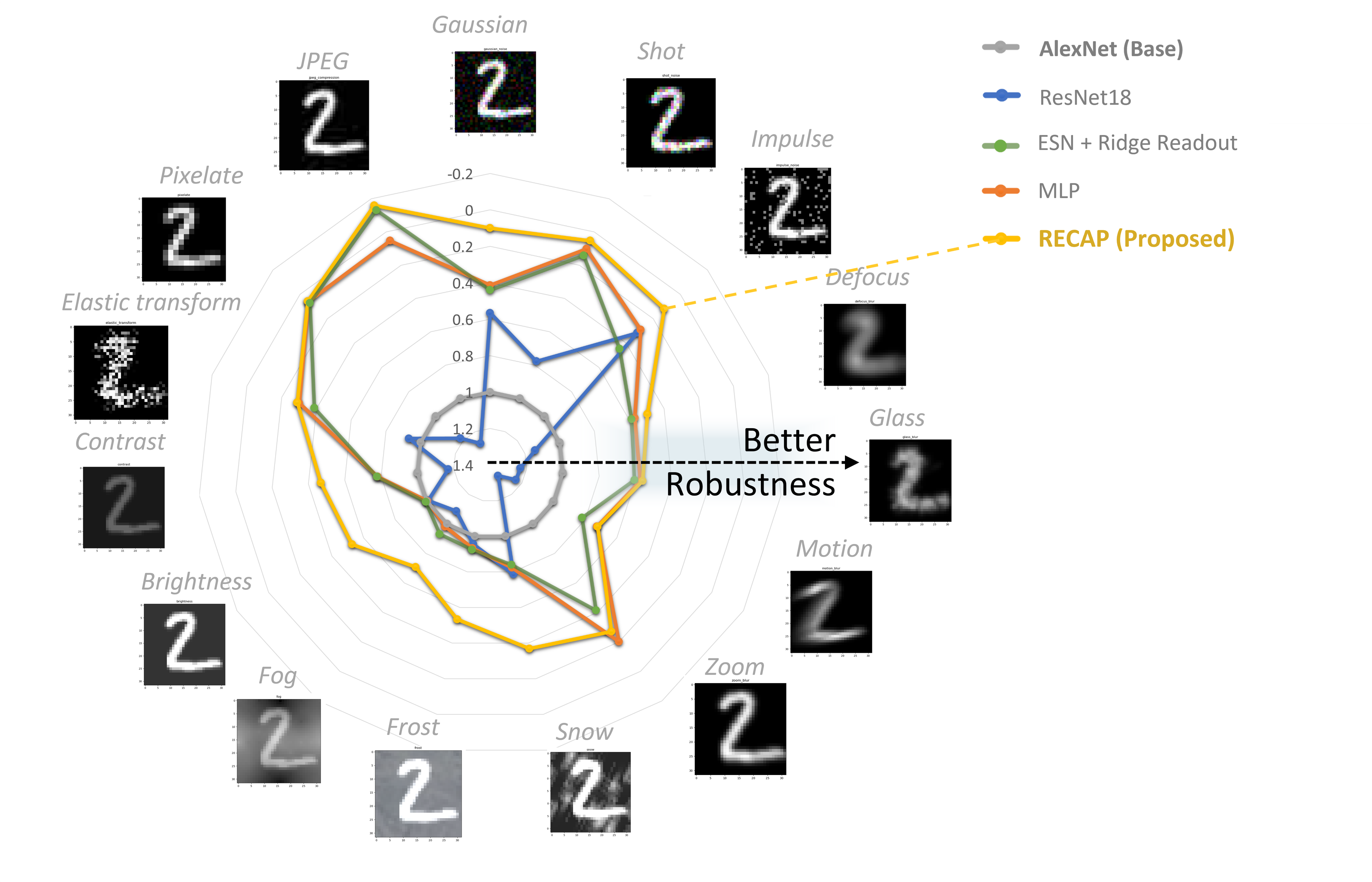}
\caption{Radar plot of per-corruption Relative CE on MNIST-C across 15 types of corruption, each represented as a point on the circular chart. The concentric contour lines indicate Relative mCE scores. For visualization, the radial axis is inverted so that lower Relative CE values (better robustness) appear farther from the center. AlexNet, used as the standard, is represented by a perfect gray circle reflecting an mCE of 100\% across all categories. The larger the area enclosed by a model's line, the better the robustness. The RECAP model, highlighted in the chart in yellow, shows the largest area, indicating superior robustness compared to baselines.}
\label{fig:radar}
\end{figure}

To visualize the \emph{shape} of robustness across corruptions, we use a radar plot (Fig.~\ref{fig:radar}) showing per-corruption Relative CE for each model.
This plot is intended to provide an at-a-glance comparison of which corruption families drive each model's robustness (or brittleness).
A full table is shown in Appendix~\ref{sec:app1} that summarizes detailed error rates across severity levels.

Notice that JPEG yields negative Relative CE for ESN-Ridge and RECAP.
This is possible under the Relative CE definition (Eq.~\eqref{eq:relce}).
A plausible explanation is that JPEG compression can suppress high-frequency variations and act like a mild denoising/regularization operator for MNIST digits, occasionally making class-discriminative strokes easier to separate.
However, we treat this as a dataset- and preprocessing-dependent phenomenon rather than evidence that compression \emph{generally} improves recognition.

\section{Discussion}
\label{sec:discussion}
The ablation results suggest that robustness improvements in RECAP arise primarily from the readout representation and local prototype updates (see Section~\ref{subsec:readout_role} for the role of each stage), rather than from training the recurrent dynamics.

\paragraph{Implications for online and continual adaptation.}
The co-activation prototype update is incremental and does not require storing full training batches or backpropagating gradients through time. This suggests that RECAP could be used in streaming settings where data arrive sequentially, with prototypes updated online. We do not evaluate continual-learning protocols in this paper, but the update structure provides a simple starting point for studying stability--plasticity trade-offs under non-stationary data.

\subsection{Robustness versus clean accuracy}
RECAP trades clean accuracy for robustness on MNIST-C.
This trade-off is not unexpected: discretization and binarization intentionally discard information, and the prototype-matching readout favors stability over fine discrimination.
More broadly, prior work has noted that robustness and standard accuracy objectives can be in tension in modern classifiers \cite{tsipras2018robustness}.
In our setting, we do not attempt to optimize the reservoir or tune the readout to maximize clean accuracy; instead, we demonstrate that a self-organizing readout rule can materially improve robustness under a clean-only training regime.
From an application standpoint, RECAP is therefore best viewed as a \emph{robustness-oriented} learning strategy rather than a replacement for fully optimized deep classifiers on clean benchmarks.

\subsection{Interpretable decision mechanism and neuro-inspired perspective}
RECAP provides an interpretable decomposition: each class $c$ is represented by a prototype $\mathbf{P}^{(c)}$, and inference is performed via overlap scoring (Eq.~\eqref{eq:score}).
This resembles a template-based decision process and connects naturally to classical ideas in associative memory and prototype recall, where stable patterns can be retrieved from noisy or partial cues \cite{hopfield1982neural,hopfield1984neurons}.
While RECAP is not a biological model, its reliance on local-style updates and population-level relational structure is aligned with neuro-inspired computation, in contrast to end-to-end gradient-trained pipelines.

\section{Limitations and Conclusion}
\label{sec:limitations}

\paragraph{Limitations.}
This mechanism-focused study is restricted to MNIST-C as a controlled testbed, and robustness behavior may differ on complex natural images. We evaluate natural corruptions only, leaving adversarial robustness and domain shift to future work. The model is bio-inspired rather than neurobiologically faithful, and a systematic sensitivity analysis across random reservoir initializations and key hyperparameters, such as the number of discretization levels $K$ and the prototype sparsity $p$, is left to future work. Also, the $N \times N$ prototype matrices may require sparse implementations to scale to larger reservoirs and more complex datasets. In addition, although the prototype update is online, we did not test sequential task learning or catastrophic forgetting protocols.

\paragraph{Conclusion.}
Despite these constraints, our results demonstrate that robustness under common corruptions can emerge from the choice of readout representation rather than corruption-specific training. By learning stable relational prototypes through Hebbian co-activation, RECAP achieves zero-shot robustness without end-to-end optimization. This offers a simple, interpretable alternative compatible with untrained dynamical systems, suggesting a promising direction for neuro-inspired robust learning.

\section*{Acknowledgments}
The author used ChatGPT and Claude to improve readability, reviewed all edits, and takes full responsibility for the content.

\bibliographystyle{elsarticle-num}
\bibliography{bib_RC_review,thesis,ircn}

\clearpage
\appendix
\section{Full result table}
\label{sec:app1}

\begin{figure}[H]
    \centering
    \rotatebox{90}{%
        \begin{minipage}{0.9\textheight}
            \centering
            \includegraphics[width=\linewidth]{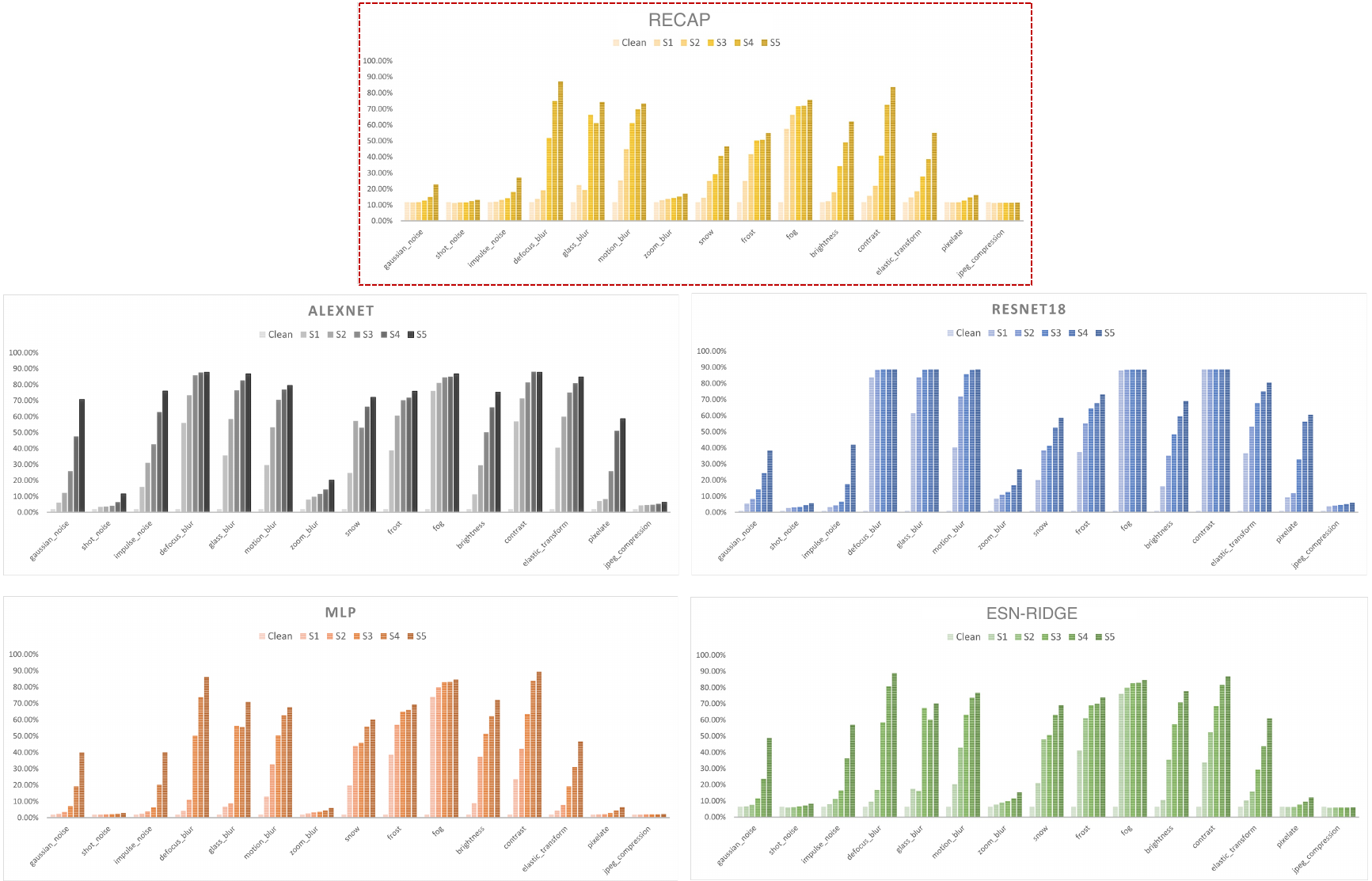}
            \caption{Detailed performances of different models on MNIST-C. The \% here is the percentage of error rate. The bar colors indicate the severity of the corruption, from light (clean) to dark (severity of 5, denoted by S5).}
            \label{fig:mnist_bar}
        \end{minipage}%
    }
\end{figure}

\end{document}